\documentclass[11pt]{article}
\usepackage[margin=1in]{geometry}
\usepackage{amsmath,amssymb}
\renewcommand{\textsuperscript}[1]{\ensuremath{{}^{#1}}}
\renewcommand{\textsubscript}[1]{\ensuremath{{}_{#1}}}
\renewcommand{\S}{\leavevmode\mathhexbox278}
\usepackage{alltt}
\usepackage{framed}
\usepackage{booktabs}
\usepackage{array}
\usepackage[hidelinks]{hyperref}
\usepackage{parskip}
\usepackage{url}
\title{How to Build Marcus's Algebraic Mind: From Minsky's Emotion-Machine Viewpoint}
\author{Hiroyuki Chuma$^{1}$ \quad Kanji Otsuka$^{2}$ \quad Yoichi Sato$^{3}$\\[0.4em]
\normalsize\itshape
$^{1}$Institute of Innovation Research, Hitotsubashi University (Professor Emeritus) \\
$^{2}$Meisei University (Professor Emeritus) \quad
$^{3}$Shuhari System}
\date{}
\begin{document}
\maketitle

\begin{framed}\small \textbf{A note on terminology.} We keep the two names distinct, but the distinction is one of \emph{capacity and incarnation}, not design-versus-implementation: both share the same architecture. \textbf{VaCoAl} (\emph{Vague Coincident Algorithm}) is that architecture --- Galois-field LFSR diffusion, block-division majority voting, reversible XOR-and-shift binding, and, as part of the base design, a collision-avoidance scheme that is \emph{endogenous} (enlarging each block's memory depth \texttt{2\textsuperscript{m}} so address conflicts become negligibly rare) and \emph{exogenous} (a dedicated multi-stage rescue circuit that intercepts the residue). Its hardware incarnation is an \textbf{SRAM-CAM (CASRAM)}, commercially realized on FPGA: fast, low-latency, and low-power, but with \textbf{small memory capacity, hence a small Frontier Size (FS)}. \textbf{PyVaCoAl} is the software incarnation of the \emph{same} architecture as a \textbf{de facto DRAM-CAM}: it runs the collision-avoidance in the host's system DRAM, so \textbf{FS can be set to gigabytes or terabytes}, and it carries design-philosophy extensions for operating at that large FS --- coping with the combinatorial explosion of large graph searches --- and for pushing the hyperdimension toward one million. PyVaCoAl is the incarnation that produced the empirical demonstrations. Two points matter for this paper. First, \textbf{FS-relative Exact Match is VaCoAl's} --- delivered by its collision-avoidance in Rescue mode --- and it is a \emph{conditional} guarantee dependent on FS, operating mode, and memory depth: a large FS (PyVaCoAl's DRAM-CAM) extends exact matching across a combinatorial frontier of tens of millions, whereas the small SRAM capacity of VaCoAl=CASRAM yields a small FS, so once the frontier exceeds that FS the exact matching it had until then can no longer be maintained. Second, \emph{faithful recovery} is therefore a property of \textbf{VaCoAl's architecture (collision-avoidance plus reversible binding)}; what \textbf{PyVaCoAl} contributes is the \textbf{large-FS DRAM-CAM realization} that let this run at the tens-of-millions scale the demonstrations required, together with \texttt{CR2} path-quality selection and multi-predicate reasoning. The nanosecond in-memory matching and the 1/N low-power rule are advantages of the SRAM-CAM incarnation, whose full hardware measurement the engine paper leaves as future work. This paper's introspection claim depends on faithful recovery within the Frontier Size, which VaCoAl's architecture provides and PyVaCoAl demonstrates at scale.\end{framed}

\begin{abstract}
This paper reports a step that turns out to have been taken already. In \emph{The Algebraic Mind}, Marcus identified three components any adequate cognitive architecture must support --- operations over variables, recursively structured representations, and a distinction between representations of individuals and of kinds --- and left the neural substrate that would carry them as an explicit conjecture. A companion paper answers that conjecture with the VaCoAl architecture, built end-to-end on XOR-and-shift over GF(2): reversible variable binding \texttt{Bind(R,F) = R $\oplus$ shift(F)}, non-commutative compositional bundling, and address-space individual/kind separation. VaCoAl's architecture also includes a collision-avoidance scheme (endogenous large memory depth plus an exogenous multi-stage rescue circuit) that makes recovery collision-free and exact within a Frontier Size; its DRAM-CAM incarnation PyVaCoAl draws that Frontier Size from system memory up to gigabytes or terabytes and so demonstrates the algebra at the scale of tens of millions of records, while its SRAM-CAM incarnation (CASRAM) trades capacity for hardware speed and low power.

Marcus's three pillars are, however, \emph{horizontal} --- they concern how a mind represents the world. This paper reads the same architecture through a second, orthogonal lens: Minsky's \emph{Emotion Machine}, whose six ascending layers supply the \emph{vertical} dimension of a mind that reasons about its own reasoning. The finding is that the vertical step required no new machinery. A representation whose structure is \emph{preserved and recoverable} is, when the structure recovered is the reasoner's own deliberative trace, an act of introspection --- Minsky's Reflective layer. Pillar 2 fixes a property of the representation and says nothing about \emph{whose} structure is recovered; point the same unbind operation inward and Reflection is what you get, cleaned by the architecture's own confidence measure. Only the argument changes.

We make three scoped claims. \emph{Capability}: exact reversible binding at O(N), held exact within a Frontier Size by VaCoAl's collision-avoidance, yields a deliberative trace that can be read back constituent by constituent with a calibrated margin --- a self-report that is faithful by construction rather than by after-the-fact verbalization. We grade deliberation into three tiers by the faithfulness of the trace it leaves --- probabilistic (LLMs), approximate-algebraic (the 2001 alternatives Marcus surveyed), and exact-algebraic (VaCoAl) --- and argue that only the third makes the reflexive read faithful rather than spurious. \emph{Necessity}: introspective audit and deliberative valuation are the same circuit at opposite settings of one parameter, and they pull in opposite directions. Repair every collision and the trace is exactly auditable while every path becomes indistinguishable from every other, so there is nothing left to credit; tolerate collisions and the \texttt{CR2} decay that credits the direct route over the detour is exactly what makes the reflexive read approximate. Credit is a record of how badly things went, and a memory that never goes badly has none to assign. \emph{Position}: we do not claim to have built the Reflective layer, and we do not claim to surpass large language models. The substrate is orthogonal to them --- it supplies the faithful, auditable trace that a probabilistic deliberation structurally cannot leave.

The same associative substrate carries two further Minsky ideas: \emph{panalogy}, which we identify with content-addressable retrieval whose best-match accuracy \texttt{CR} is the barometer of analogical fit and whose one-shot, real-time, accumulating writes (the Kanerva SDM lineage) give continual, lifetime learning and transfer to novelty; and \emph{credit assignment}, which we show is reflexive counterfactual simulation --- surgically unbinding an earlier choice, rebinding a contrary-to-fact one, and comparing --- so that the introspective (Minsky) and causal (Pearl) axes interlock, both requiring exact reversibility. Where the companion climbs Pearl's causal ladder off the same reversibility, this paper climbs Minsky's introspective stack and locates the point at which the two meet.

\emph{Not claimed.} Reflection's \emph{sufficient} conditions --- a meta-control loop and temporal context-tagging --- are named as future work, not supplied. Reflection is read strictly as introspection, not evaluation; evaluative self-assessment requires value binding the architecture does not have. Object-level deliberation is \emph{demonstrated} by PyVaCoAl; Reflection's reachability is \emph{argued}. And every speed and power figure belongs to the SRAM-CAM incarnation and is, in this paper, unbenchmarked. Appendix A grades every claim in the paper by which of these categories carries it.

\end{abstract}

\section{Introduction}
\subsection{Two closings, one substrate}
\emph{The Algebraic Mind} closed with an open question: even granting that symbol manipulation rests on operations over variables, structured representations, and records for individuals, it remained to be discovered how those components are implemented in neural hardware. Marcus offered a candidate --- registers arranged into \emph{treelets}, built by cascading developmental programs rather than by gradient descent --- and marked it plausible but speculative. The companion paper [Chuma et al., 2026b] takes the position that the substrate the conjecture called for is now available: the VaCoAl architecture, organized around a single elementary primitive, XOR-and-shift over GF(2), realized by primitive-polynomial linear-feedback shift registers. It maps the three pillars, component for component, onto VaCoAl's operational commitments, and reinterprets the treelet as an algebraic register set whose identity is a primitive generator polynomial.

That correspondence is complete at the level Marcus pitched it, which is \emph{horizontal}: all three pillars concern how a mind represents the world --- its variables, its structures, its individuals. None of them concerns how a mind represents \emph{its own thinking}. For that, a second framework is needed, orthogonal to Marcus's. Minsky's \emph{Emotion Machine} supplies it: six ascending layers --- Instinctive Reactions, Learned Reactions, Deliberative Thinking, Reflective Thinking, Self-Reflective Thinking, Self-Conscious Emotions --- each observing and reasoning over descriptions of the layer beneath it. Where Marcus specifies \emph{what} an algebraic mind must be able to do, Minsky specifies \emph{how the doing is stacked}, and in particular what it takes to climb from deliberating about the world to introspecting on one's own deliberation.

\subsection{The hinge}
The two pictures meet at one point, and it is already inside the companion's second pillar. Pillar 2 requires that a structured representation keep its constituents and their arrangement \emph{recoverable}. The requirement fixes a property of the representation; it does not fix \emph{whose} structure is recovered. Point the recovery at the world and you get object-level structured cognition --- Minsky's Deliberative layer. Point it at \emph{the record of one's own most recent deliberation} and ``recover the constituents and their arrangement'' becomes ``recover what I just bound to what, and how'' --- which is introspection, Minsky's Reflective layer. The vertical step from Deliberation to Reflection is the horizontal Pillar-2 requirement turned inward. This paper develops that hinge, and shows that only an \emph{exact} recoverability turns it safely.

It is worth registering how little this costs, because the smallness is the point. Verticality is usually treated as an addition: a monitor placed above the reasoner, a second model of the first, a module that watches. On the present reading it is a redirection. The same operation, the same algebra, the same cost --- one changed argument. If that is right, then the price of a mind that can watch itself think was paid in full when the mind was given a structure it could take apart, and everything after is control rather than capability.

\subsection{Why introspection is a record of roads not taken}
Before the architecture, one illustration, because the property this paper turns on is easy to mistake for bookkeeping.

Consider what Minsky's Reflective question actually asks for. \emph{How did she choose which resources to use?} The answer is not the choice. It is the choice together with what it was chosen \emph{over} --- the method adopted, the evidence it rested on, and the alternative set aside. Strip the alternative out and the remaining report is not an account of a decision at all; it is an account of an outcome, and an outcome is compatible with there having been no deliberation whatever. The stored object that makes introspection possible is a record of tried and abandoned branches, each with something attached saying how it went.

This is precisely what a probabilistic deliberation does not leave behind. A transformer's context is, by construction, the surviving sequence alone: the candidates discarded during sampling leave no trace, and lengthening the context window multiplies tokens without adding a single branch. The chain of thought that can be read afterwards is a re-serialization, and it need not match the computation that produced it [Turpin et al., 2023]. The difference between that and a deliberative trace is one of type, not of size --- more capacity yields a longer surviving sequence, never a rejected one.

Two structural properties are required to store such a record, and the architecture supplies both. \emph{Reversibility}: because \texttt{Bind(R,F) = R $\oplus$ shift(F)} is exactly invertible, one can unbind at a role, inspect what occupies it, and put something else there, so a branch that was abandoned can be revisited rather than merely regretted. \emph{Determinism}: because the diffusion is algebraic rather than sampled, the same branch is bit-exactly reproducible, so two roads can be compared on the same footing rather than recalled through the fog of a second sampling. Section 7.4 shows that this is exactly what credit assignment needs, and \S{}8 shows what the architecture must give up to have it.

\subsection{What is and is not claimed}
Because this paper joins an engineering substrate to a cognitive-architecture framework written without reference to hardware, it is worth stating at the outset which claims are load-bearing and which are not.

\textbf{Capability.} Exact reversible binding, held exact within a Frontier Size by VaCoAl's collision-avoidance, yields a deliberative trace whose constituents can be recovered individually, in their roles, each with a margin the architecture reports rather than estimates. We claim this makes a self-report faithful \emph{by construction}: the system reports what it bound because the report is a read of the binding, not a narration produced alongside it. We claim further that this combination is structurally unavailable to substrates whose recovery is approximate, which can return a nearest candidate but cannot guarantee it is the one deliberated.

\textbf{Necessity.} We claim that exact-algebraic deliberation supplies the \emph{necessary substrate} for faithful introspection, and thereby brings Minsky's Reflective layer into reach. We claim in addition, and this is the sharper form, that the architecture's two Minsky-relevant functions --- introspective audit and deliberative valuation --- are not two features but one circuit at two settings, and that they pull in opposite directions. A memory that repairs every failure can be audited exactly and has nothing to credit; a memory that tolerates failure can credit its own routes and audits them only approximately. This is developed in \S{}8, and it cuts against the reflex of treating error suppression as an unqualified good.

\textbf{Position.} We do not claim to have built the Reflective layer, and we do not claim to surpass large language models on any task they are for. The layer this substrate supplies is orthogonal to theirs: a faithful, factorizable trace, which a probabilistic deliberation structurally cannot leave, and on which a reflective controller could be built.

\textbf{Not claimed.} Reflection's \emph{sufficient} conditions --- a control loop that decides when to introspect, and a tagging scheme that separates past thought from present --- are named as future work (\S{}10), not supplied here. We read Reflection strictly as \emph{introspection} (descriptive recovery of one's own deliberative structure), not as \emph{evaluation} (judging whether that deliberation was good or principled), which belongs to the Self-Reflective and Self-Conscious layers above and requires value binding the architecture does not supply. Object-level deliberation is \emph{demonstrated} by PyVaCoAl; Reflection's reachability is \emph{argued} here. No introspection experiment is reported: the reflexive-read predictions of \S{}10 are tests, not results. And the nanosecond and low-power figures belong to the SRAM-CAM incarnation and are unbenchmarked in this paper. Appendix A tabulates which category carries which claim, so that a reader can check the ledger rather than take our word for the labels.

\subsection{Relation to the companion papers, and contributions}
Three sibling papers precede this one: the engine paper [Chuma et al., 2026a, arXiv:2604.11665], which develops the VaCoAl architecture and its two incarnations --- SRAM-CAM (CASRAM) and the DRAM-CAM PyVaCoAl; the Algebraic-Mind paper [Chuma et al., 2026b, arXiv:2605.21379], which maps Marcus's three pillars onto the engine and climbs Pearl's causal ladder as an extension; and the hippocampal Perspective [arXiv:2605.15652], which reads the DG--CA3 circuit as a biological homologue of the engine. This paper is the fourth vantage. Its contributions:

\begin{enumerate}
\item It adds the Emotion-Machine viewpoint --- the \emph{vertical}, self-directed axis the three pillars lack --- and maps the architecture onto Minsky's six layers (\S{}5).
\item Its core claim (\S{}6): Pillar-2 recoverability applied reflexively is Minsky's Reflection, realized as reflexive unbinding of one's own role--filler bundle under VaCoAl's confidence measure.
\item It grades deliberation into three tiers by trace faithfulness (\S{}4), placing LLMs, the 2001 alternatives, and VaCoAl on one axis, and showing that only exact reversibility makes the reflexive read faithful.
\item It distinguishes the two vertical axes the same reversibility unlocks --- Pearl's causal ladder (the companion) and Minsky's introspective stack (this paper) --- and locates deliberative valuation and introspective audit at two settings of a single reversibility knob (\S{}8).
\item It identifies Minsky's \emph{panalogy} with content-addressable retrieval (so \texttt{CR} is the graded barometer of analogical fit, and one-shot/real-time/continual learning follows from the SDM lineage), and shows Minsky's \emph{credit assignment} to be reflexive counterfactual simulation --- the point at which the causal and introspective axes interlock, both requiring exact reversibility (\S{}7).
\item It states, as a constraint rather than a configuration detail, that audit and valuation cannot be maximized together (\S{}8), and gives in Appendix A a ledger of which of this paper's claims are demonstrated, which architectural, which argued, and which deferred.
\end{enumerate}

\subsection{Structure of the paper}
Section 2 summarizes the architecture, the pillars as the companion answers them, and Minsky's stack. Section 3 separates the two vertical axes that rise from association and says where they interlock. Section 4 grades deliberation into three tiers by trace faithfulness. Section 5 maps the six layers onto the substrate. Section 6 is the hinge: Reflection as reflexive unbinding. Section 7 develops panalogy, one-shot transfer, continual learning, and credit assignment on the same associative memory. Section 8 is the paper's second core: audit and valuation as one knob at two settings, and why they cannot both be maximized. Section 9 notes two independent corroborations and the boundaries that keep them honest. Section 10 gives reservations, the chief open problem, and testable predictions. Section 11 concludes by carrying two consequences out of the paper separately from the architecture that produced them. Appendix A is the claim ledger.

\section{Background}
\subsection{The VaCoAl architecture, and its SRAM-CAM and DRAM-CAM incarnations}
We summarize the architecture at the level needed here; full development is in [Chuma et al., 2026a,b]. Commitments 1--3, the binding algebra, and the collision-avoidance scheme below are VaCoAl's; \S{}2.1's closing paragraph states how the DRAM-CAM incarnation PyVaCoAl scales them from system memory and what that scaling means for this paper's claim.

\emph{Commitment 1 --- one primitive.} Every operation --- diffusion, binding, unbinding, bundling --- decomposes into XOR-and-shift over GF(2). XOR is the field's addition and is its own inverse; shift (multiplication by \texttt{x mod G(x)}) supplies non-commutativity.

\emph{Commitment 2 --- deterministic reversible diffusion.} For a primitive polynomial \texttt{G(x)} of degree \texttt{m}, diffusion \texttt{$\Psi$(P) = x\textsuperscript{m}$\cdot$P(x) mod G(x)} runs on a linear-feedback shift register; it is GF(2)-linear, bijective off the collision kernel, and satisfies a quasi-orthogonal-diffusion (QOD) property: distinct inputs produce outputs whose Hamming distance concentrates near \texttt{m/2} with variance $\approx$ \texttt{m/4}, exponentially tight in \texttt{m}. A one-bit input change flips about half the output bits --- the deterministic Avalanche Effect that replaces Kanerva's stochastic random projection.

\emph{Commitment 3 --- voting readout with confidence.} A length-\texttt{L} hypervector is split into \texttt{N} blocks, each under its own primitive polynomial. Readout is block-wise majority voting, yielding a per-step Confidence Ratio \texttt{CR1 = (winner votes)/N} and a path-integral \texttt{CR2(n) = $\Pi$\textsubscript{i} CR1(i)}.

\emph{Binding and unbinding.} \texttt{Bind(R,F) = R $\oplus$ shift(F)}; because XOR self-inverts and shift is bijective, \texttt{Unbind(B,R) = shift\textsuperscript{-1}(B $\oplus$ R) = F} recovers the filler exactly in the noiseless case, at O(N) cost. A bundle \texttt{Repr = $\oplus$\textsubscript{i} Bind(R\textsubscript{i},F\textsubscript{i})} is a hypervector in the same space, so composition is fixed-dimensional and recursive; querying \texttt{Repr} with a role returns the corresponding filler plus crosstalk that the QOD property scatters and the voting readout removes.

The three commitments are not independent design decisions taken in sequence. Commitment 1 fixes the algebra; Commitment 2 is what that algebra does when iterated inside a block; Commitment 3 is what a bank of such blocks does when their outputs are counted. There is one decision in this architecture, and the rest is arithmetic --- which is why, in \S{}6, the reflexive turn needs no fourth commitment.

\emph{What VaCoAl provides, and how PyVaCoAl scales it.} The properties above are VaCoAl's architecture, and the noiseless identity \texttt{Unbind(Bind(R,F),R) = F} is exact by GF(2) algebra alone. At a fixed memory depth, address collisions begin to appear once stored records exceed the maximally available SRAM, and LFSR diffusion \emph{alone} can no longer maintain the orthogonality the majority vote needs; within SRAM capacity (e.g., 8-24Mb for Intel Arria V GX), VaCoAl forestalls this with its collision-avoidance scheme --- endogenous (enlarging the memory depth \texttt{2\textsuperscript{m}}) and exogenous (the multi-stage rescue circuit) --- which is part of the base architecture, not an afterthought, and which both incarnations carry. The incarnations differ in \emph{capacity}. PyVaCoAl realizes the architecture as a de facto DRAM-CAM, so the Frontier Size can be drawn from the host's system memory up to gigabytes or terabytes, and it is this large FS (e.g., with the decade-old workstation with 128GB RAM) that let PyVaCoAl run the algebra at scale: on a $\sim$470,000-edge WIKIDATA mentor--student ontology it traced the mentor genealogies of all 64 Fields Medalists back as far as 57 generations, over 25.5 million paths, synthesizing composite concepts by bundling and probing them by unbinding under the \texttt{CR} score, with collisions driven to zero in Rescue mode. Two operating modes follow, both VaCoAl's: in \textbf{Rescue mode} the exogenous circuit suppresses collisions to zero and attains Exact Match relative to the Frontier Size; in \textbf{Don't Care mode} minute collisions are tolerated in exchange for \texttt{CR2}-based semantic selection. FS-relative Exact Match is a conditional guarantee (dependent on FS, mode, and memory depth), so PyVaCoAl's large DRAM-borne FS extends it across a tens-of-millions frontier, whereas the small SRAM capacity of VaCoAl=CASRAM yields a small FS beyond which exact matching cannot be sustained. The upshot for this paper: \emph{reversible binding and collision-free recovery within the Frontier Size} are VaCoAl's architecture; the \emph{scale} at which they were demonstrated is PyVaCoAl's large-FS DRAM-CAM realization. The nanosecond matching and 1/N low-power rule belong to the SRAM-CAM incarnation.

\subsection{Marcus's three pillars, as answered by the companion}
\emph{Pillar 1 (operations over variables).} An operation defined uniformly over every instantiation of a variable, generalizing freely to unseen instances (Marcus's universally quantified one-to-one mappings). Answered by reversible XOR binding: the same role binds any filler, seen or unseen, with the same algebraic effect, and unbinding recovers it --- free generalization by construction, not by training coverage.

\emph{Pillar 2 (structured representations).} Complex wholes whose constituents and their arrangement are preserved and jointly recoverable; the diagnostic is that identical constituents in reversed roles must remain distinguishable and recoverable. Answered by non-commutative compositional bundling: \texttt{Bind(R,F)} differs from \texttt{Bind(F,R)} because shift acts asymmetrically, and unbinding is bit-exact rather than the accumulating quasi-inverse of circular convolution.

\emph{Pillar 3 (individuals vs kinds).} A persistent per-individual identifier distinct from the individual's current properties and from the kind it instantiates. Answered by Entry Addresses drawn from a quasi-orthogonal space, an \texttt{ISA} role binding the individual to its kind, and a separate property bundle --- new individuals added by drawing a fresh address, without retraining.

\subsection{Minsky's Model Six}
Minsky's stack is not a monolith: each layer reasons over descriptions of the layer below. The layer of interest is the fourth. In his worked example, the question ``how did she choose which resources to use'' is Reflective; ``did she make a good decision'' is Self-Reflective; ``did her actions live up to her principles'' is Self-Conscious. Reflection is therefore \emph{introspective} --- descriptive access to the structure of one's own recent thinking --- and not yet \emph{evaluative}. Minsky also notes that reflection works over short-term \emph{records} of what the lower levels did, and must track the \emph{context} in which those records were made, so that past thought is not confused with present thought. Both features --- a readable record and a context tag --- reappear below as the architecture's bundle and as the paper's chief open problem.

\section{Two Ways Up From Association}
Association --- retrieving what tends to co-occur --- is the floor. Two distinct vertical axes rise from it, and the same property of the architecture, exact reversibility, unlocks both.

The companion climbs the first: \textbf{Pearl's causal ladder}. Rung-2 intervention (the do-operator) is supported operationally by surgical unbind/rebind --- recover a role's filler, substitute a counterfactual filler, rebind --- a sequence of XOR-and-shift steps that disturbs only the targeted binding, where a lossy quasi-inverse would inject noise from every other binding. Rung-3 counterfactuals, requiring simultaneous factual and contrary-to-fact worlds, are argued to rest on the two-orthogonalizer architecture with \texttt{CR2} path traces as estimator. This axis is about causal reasoning over the \emph{world}.

This paper climbs the second: \textbf{Minsky's introspective stack}. Its steps are not about the world but about the \emph{self} --- a mind reasoning over descriptions of its own lower-level activity. Pearl asks what would happen under intervention; Minsky asks how the reasoner just reasoned. The two axes begin apart, and both are gated by the same reversibility: surgical recoverability for Pearl, faithful trace recoverability for Minsky. An architecture whose composition is irreversible can climb neither. But they are not merely parallel: they \textbf{interlock} at credit assignment (\S{}7). Assigning credit to one's own earlier choices --- Minsky's engine of learning-how-to-think --- means counterfactually altering a past choice and comparing outcomes, which is Pearl's intervention/counterfactual turned reflexively on the self's own trace. The companion establishes the causal climb; we develop the introspective one and show where the two meet.

The word \emph{gated} is doing precise work and we mean it strictly. Neither axis is a matter of doing something slightly better on a lossy substrate. An intervention that perturbs the bindings it was supposed to hold fixed is not a less precise intervention; it is a different operation. A self-read that returns a filler near the one deliberated is not a slightly noisy self-report; it is a report about something else. In both cases the failure is of type rather than of degree, which is why neither axis is a frontier that a lossy substrate crosses by getting larger.

\section{Three Tiers of Deliberation}
We grade deliberation by the faithfulness of the trace it leaves --- whether one can later recover \emph{what was composed into what}, and whether that recovery is guaranteed correct.

\textbf{Tier 1 --- probabilistic (LLMs; eliminative connectionism).} Composition is soft and non-invertible; the only record is an after-the-fact re-serialization into text, often unfaithful to the underlying computation [Turpin et al., 2023]. This tier fails all three of Marcus's pillars --- no operation over variables, no recoverable structure, no stable records for individuals --- and so has no trace to introspect, only a verbal shadow.

\textbf{Tier 2 --- approximate-algebraic (the 2001 alternatives).} Tensor products, holographic reduced representations, and temporal-synchrony schemes compose algebraically and answer Pillar 2 \emph{partially} --- exactly the partial answers Marcus catalogued and the companion \S{}10 dispatches. Tensor products preserve structure but explode dimensionally (a five-level embedding on the order of 10\textsuperscript{7} nodes). Circular convolution keeps dimension fixed but replaces exact unbind with a quasi-inverse whose crosstalk accumulates \emph{in the signal subspace}, limiting recursion depth. Recent representational-geometry analysis makes the failure sharp: factorization recovery succeeds only for large dimension and low correlation, and converges to \emph{spurious} factorizations when correlations are high [Poduval et al., 2026]. The trace is recoverable, but approximately and conditionally, and it can be silently wrong.

\textbf{Tier 3 --- exact-algebraic (VaCoAl).} Binding is exactly invertible and retrieval is deterministic voting with an explicit confidence; collection recovery is made exact within a bounded Frontier Size by VaCoAl's collision-avoidance in Rescue mode --- realized at the tens-of-millions scale of the demonstrations by PyVaCoAl's large-FS DRAM-CAM. The trace is faithful by construction, within the Frontier Size.

\begin{center}\small
\begin{tabular}{@{}p{2.9cm}p{3.5cm}p{4.4cm}p{4.4cm}@{}}
\toprule
\textbf{Tier} & \textbf{Composition} & \textbf{What a self-read returns} & \textbf{Characteristic failure} \\
\midrule
1. Probabilistic (LLM) & Soft, non-invertible mixture & Nothing; a re-serialization produced alongside the computation & Unfaithful account, undetectable from inside \\
\addlinespace
2. Approximate-algebraic & Algebraic, quasi-invertible & The filler plus crosstalk accumulating in the signal subspace & Silent settling on a constituent never deliberated \\
\addlinespace
3. Exact-algebraic (VaCoAl) & Algebraic, exactly invertible & The bound filler, with the voting margin \texttt{CR1} & Collision saturation past the Frontier Size, reported not hidden \\
\bottomrule
\end{tabular}
\end{center}

\begin{center}\small\emph{Table 1. Three tiers of deliberation, graded by the faithfulness of the trace. The
rightmost column is the one that matters for Reflection: what goes wrong, and whether the
system can tell.}\end{center}

The tiers are ordered by exactly the property introspection needs. Tier 1 has no trace to read. Tier 2 has a trace whose read can return a false constituent --- the vector-symbolic analogue of unfaithful chain-of-thought. Tier 3 has a trace whose read is guaranteed correct within its frontier.

Note where the ordering does \emph{not} run. It is not an ordering by expressive power, by fluency, or by usefulness; on all of those Tier 1 is plainly ahead, and nothing in this paper disputes that. It is an ordering by one property, and the property is legible only when the question asked is about the system's own reasoning rather than about the world. This is the sense in which the substrate is orthogonal to the language model rather than in competition with it: they are ranked on different axes, and each is first on its own.

\section{Minsky's Layers on the Substrate}
The six layers map onto the architecture as follows.

\emph{Instinctive (L1) and Learned Reactions (L2).} Direct and acquired stimulus--response: single-step associative retrieval, an Entry-Address lookup cleaned by voting, reported with \texttt{CR1}. L2 differs only in that the bindings were acquired rather than innate.

\emph{Deliberative Thinking (L3).} Multi-hop reasoning over alternatives: iterated bind/unbind chains. This is the layer PyVaCoAl exercises directly --- the mentor-genealogy traced across 57 generations and 25.5 million paths, and the soap-opera role--filler cycle used in the companion as the Pillar-2 diagnostic. The \texttt{CR2} path integral is the natural deliberation-quality measure, accumulating confidence along the reasoning chain.

\emph{Reflective Thinking (L4).} Introspection --- the hinge developed in \S{}6. The same unbind operation, pointed at the reasoner's own L3 trace.

\emph{Self-Reflective (L5) and Self-Conscious (L6).} Evaluation and normative appraisal --- judging whether the deliberation was good or principled --- requiring the binding of value and affect. Above this paper's scope (\S{}10).

\begin{center}\small
\begin{tabular}{@{}p{3.4cm}p{5.2cm}p{5.2cm}@{}}
\toprule
\textbf{Minsky layer} & \textbf{Operation on the substrate} & \textbf{Status in this paper} \\
\midrule
L1 Instinctive & Entry-Address lookup, voting readout & Demonstrated (single-step retrieval) \\
L2 Learned reactions & The same, over acquired bindings & Demonstrated \\
L3 Deliberative & Iterated bind/unbind chains; \texttt{CR2} along the path & Demonstrated (PyVaCoAl, 57 generations) \\
L4 Reflective & The same unbind, argument turned inward (\S{}6) & Argued reachable; sufficient conditions deferred \\
L5 Self-Reflective & Requires value binding & Out of scope \\
L6 Self-Conscious & Requires norm and affect binding & Out of scope \\
\bottomrule
\end{tabular}
\end{center}

\begin{center}\small\emph{Table 2. The six layers on the substrate, with the status of each claim. The
break between L3 and L4 is the hinge; the break between L4 and L5 is the boundary of this
paper.}\end{center}

The division is clean: L1--L3 are world-directed and are the companion's territory; L4 is the reflexive turn; L5--L6 need machinery the architecture does not yet supply.

\section{Reflection as Reflexive Unbinding}
Here is the paper's core. At the object level, Pillar 2 stores a structured thought as a bundle of role--filler bindings and recovers its parts by unbinding. Consider a deliberative trace --- the record of an L3 deliberation about, say, which resource to use for a goal:

\begin{alltt}
D = Bind(Goal, g) \(\oplus\) Bind(Method, m) \(\oplus\) Bind(Evidence, e) \(\oplus\) Bind(Alternative, a)
\end{alltt}

This is an ordinary Pillar-2 bundle; its roles happen to be the reasoner's \emph{own deliberative roles}. Introspection --- Minsky's ``how did she choose which resources to use'' --- is then nothing more than querying \texttt{D} with those roles:

\begin{alltt}
Unbind(D, Method)      = m          "the method I used"
Unbind(D, Alternative) = a          "the option I rejected"
Unbind(D, Evidence)    = e          "what I based it on"
\end{alltt}

each returned with a confidence \texttt{CR1}, and the introspective chain accumulating \texttt{CR2}. No new machinery is introduced: introspection is Pillar-2 unbinding with the trace pointed at the self. This is the precise sense in which the companion's second pillar \emph{already contains} Minsky's Reflective layer --- the operation was there all along; only its argument is turned inward.

The middle line is the one to dwell on. \texttt{Unbind(D, Alternative) = a} recovers a road not taken, and it recovers it because the road was bound into the trace as a constituent rather than discarded when the decision resolved. This is the concrete form of \S{}1.3: what makes the report an account of a \emph{decision} is that the rejected branch is still addressable. An architecture that stores only what it settled on can answer ``what did I conclude'' and cannot answer ``how did I choose,'' and the second is the whole of Minsky's Reflective question.

\textbf{Why the tier matters.} Reflexive recovery inherits the reliability of the recovery operation, and this is where the three tiers separate decisively. In Tier 1 there is no \texttt{D} to query --- only a post-hoc verbalization that need not match the computation. In Tier 2 the unbind returns \texttt{m $\oplus$ $\xi$}, where \texttt{$\xi$} is crosstalk that accumulates \emph{in the same subspace as the signal}; under high load the readout can settle on a filler \texttt{m'} that was never the one deliberated. A system introspecting through such a recovery would report ``I used \texttt{m}'' when its trace best-matches \texttt{m'} --- a false account of its own reasoning, the vector-symbolic form of unfaithful chain-of-thought, and precisely the failure a Reflective layer must not have. In Tier 3 the QOD property scatters \texttt{$\xi$} across the address space rather than accumulating it, and majority voting contracts to the true filler, so \texttt{Unbind(D, Method) = m} exactly within the frontier, with \texttt{CR1} reporting the margin. The noiseless exactness is VaCoAl's algebra; the guarantee that it \emph{holds at scale} --- that collisions do not silently corrupt the reflexive read once the trace store grows past the few-hundred-thousand-record point where conflicts begin --- is VaCoAl's collision-avoidance architecture, realized across a tens-of-millions frontier by PyVaCoAl's large-FS DRAM-CAM, and holding on the SRAM-CAM incarnation up to the smaller Frontier Size its capacity affords. The self-report is faithful by construction within that frontier, and its confidence is auditable. Marcus's second pillar, met \emph{exactly} and held exact within the Frontier Size, is what makes the reflexive read safe.

The general form of that paragraph is worth stating once, because it is not about our substrate. A representation that cannot be exactly taken apart was never composed; it was mixed. Introspection is decomposition pointed at oneself, so a mind whose composition is lossy does not introspect a little less accurately --- it introspects something that was never separately there. The corollary is the uncomfortable one: an architecture that lacks this cannot acquire it by gaining a monitoring component, because exactness of inverse is a property of the elementary operation and not of what is stacked on it. Reflection cannot be bolted on.

\textbf{Two further requirements, two further mechanisms.} Minsky's Reflection must report \emph{how sure} its self-read is; \texttt{CR1} is that report, a built-in margin rather than a post-hoc estimate --- notably, an \emph{under}-confidence signal (a low \texttt{CR1}) where deep networks would be silently overconfident. And it must be able to \emph{decompose} a composite thought on demand; exact unbind is that decomposition, single-step, where Tier 2 would need an iterative, convergence-conditional resonator.

\textbf{Reflection is introspection, not evaluation.} The claim is bounded by Minsky's own layer tags. Recovering \texttt{m} and \texttt{a} answers \emph{how} the reasoner chose and what it set aside; it does not answer whether the choice was \emph{good} (Self-Reflective) or \emph{principled} (Self-Conscious). Introspection reduces to faithful recovery of the trace's structure --- faithful factorization of one's own deliberation --- and needs no value or norm system. Evaluation needs the binding of value and affect, which the architecture does not supply, and which we therefore assign to the layers above. Reading Reflection as introspection keeps the necessary condition at ``faithful, factorizable trace,'' with no evaluative machinery smuggled in.

\section{Panalogy, Associative Memory, and Credit Assignment}
Introspection (\S{}6) reads a stored trace. Two of Minsky's central ideas --- \emph{panalogy} and \emph{credit assignment} --- show that the same associative substrate does more than store: it is what lets a mind transfer to novel situations and learn which of its own choices deserve credit. Both rest on the substrate being a content-addressable (associative) memory, and both are graded by the same \texttt{CR} measure.

\subsection{Panalogy is content-addressable retrieval}
Minsky's \emph{panalogy} (``parallel analogy'') represents one thing simultaneously across several realms --- physical, possessional, mental --- with corresponding features tied to the \emph{same slots} of one structure, so that a cue in one realm reaches the linked representations in the others and the mind switches realms almost instantly. In the algebra this is a bundle of parallel role-bindings over shared roles, \texttt{Panalogy = Bind(Physical, $\cdot$) $\oplus$ Bind(Possession, $\cdot$) $\oplus$ Bind(Mental, $\cdot$)}, the shared ``slots'' being shared role vectors. Retrieving the analogue in another realm from a partial cue is exactly \emph{content-addressable, best-match} retrieval --- which is what the substrate is: the engine paper's core contribution is an address decoder that preserves the similarity (best-match) principle of Kanerva's SDM while scaling past the kilobit ceiling of brute-force content-addressable memory. Two consequences follow. First, the \emph{accuracy} of the associative memory (does best-match return the right analogue?) and its \emph{comprehensiveness} (does it hold all the linked realms at scale?) are the barometer of panalogy's accuracy and coverage. Second, \texttt{CR} --- the best-match margin from majority voting --- is the graded measure of \emph{how well} a situation fits a stored panalogy. Panalogy read as graded similarity is thus exact once it is seen as a CAM retrieval: \texttt{CR} is its dial.

\subsection{Novelty, change, and anomaly: one-shot analogical transfer}
Because retrieval is best-match rather than exact-key, a \emph{novel} problem retrieves its nearest stored analogues --- panalogy-based transfer to a situation never seen. This is the substrate's answer to change and anomaly: an unfamiliar input is met by the closest familiar structure, graded by \texttt{CR}, rather than by failure. It is also why the learning is one- or few-shot: associative, distributed memory recognizes from one or a few examples, a general property of HDC (Rahimi et al., 2018; Mitrokhin et al., 2019; Poduval et al., 2026) inherited by VaCoAl as a member of the Kanerva SDM lineage and made lossless by its exact CAM. Minsky names this faculty directly: in the credit-assignment section of \emph{The Emotion Machine} he attributes skill at transfer of learning to other realms to skill at making and using panalogies. Object-level transfer to novelty is panalogy retrieval over a content-addressable memory.

\subsection{Real-time, continual, lifetime learning}
Storage in this lineage accumulates by \emph{writing}, in real time, without gradient descent and without disturbing prior records: a new individual is a fresh Entry Address deposited in the CAM, leaving every existing record intact (the companion's Pillar 3). Memory therefore grows online --- continual, lifetime learning by construction, as in Kanerva's SDM. VaCoAl inherits this, and its collision-avoidance --- realized at the tens-of-millions scale by PyVaCoAl's large-FS DRAM-CAM --- keeps the accumulating store collision-free and exactly recoverable within the Frontier Size, so continual accumulation does not degrade into interference. Real-time, one-shot, and continual are not features bolted on; they are what an associative memory that writes once and reads by content simply \emph{is}.

\subsection{Credit assignment as reflexive counterfactual}
Minsky's keyword for how a mind learns which of its resources to trust is \emph{credit assignment} --- in his words, a memory record of how one solved a problem --- and he is explicit that its bottleneck is introspection: it is harder to see \emph{how} one found a solution than to find it, and our ability to introspect is limited. That is exactly the gap the faithful reflexive read of \S{}6 fills. But credit assignment is more than reading the trace: Minsky says to credit not the final act, nor even the strategy, but the \emph{earlier choices that selected the winning strategy}, and that one can dwell on what might have led to a success. That is counterfactual attribution over one's own trace: hold the factual deliberation \texttt{D}, surgically unbind an earlier choice, substitute a contrary-to-fact filler, rebind, re-run, and credit the choice by the difference. Structurally this is the companion's Pearl rung-2/3 machinery --- surgical intervention and parallel factual/counterfactual worlds --- turned \emph{reflexively} on the self. It is therefore the point at which the two vertical axes of \S{}3 stop being parallel and interlock: reflective credit assignment \emph{is} counterfactual simulation applied to one's own deliberative trace, and both demand the same reversibility, which lossy substrates lack. And the path measure is the credit signal: \texttt{CR2}, the product of per-hop confidences along the associative path, distributes credit along the reasoning chain, so the Don't-Care-mode \texttt{CR2} decay that prunes circuitous routes and keeps direct ones (\S{}8) is credit assignment in action --- crediting the direct path, discrediting the detour.

That last observation has a consequence the next section is entirely about, and it is not a comfortable one. If \texttt{CR2} is the credit signal, then credit is manufactured out of the small failures the memory tolerated along a route. A route that never lost a block accrues no distinguishing mark. Credit assignment therefore does not sit on top of exact recall; it sits on top of \emph{inexact} recall, and the two Minsky functions this paper has developed --- audit in \S{}6 and credit in \S{}7.4 --- turn out to want the architecture set two different ways.

The loop closes on Minsky's own claim. Better credit assignment yields better transfer (\S{}7.2); transfer is panalogy retrieval (\S{}7.1); panalogy is content-addressable; and credit assignment is reflexive counterfactual over the same content-addressable trace. Object-level one-shot adaptation to novelty and meta-level reflective credit assignment are the two faces of one reversible associative memory --- which is why an \emph{exact} CAM, not merely a fast one, is what building this part of Marcus's mind requires.

\section{No Failure, No Credit: Two Settings of One Knob}
The exactness that makes reflexive recovery faithful does not preclude selection; it relocates it. And the relocation is not free, which is the finding of this section and the second thing we would like carried out of this paper.

\subsection{The transition}
Poduval et al.'s \textbf{encoder-width knob \texttt{w}} is a representational-geometry control on a \emph{lossy} substrate: small \texttt{w} gives near-orthogonal codes good for decoding and factorization, intermediate \texttt{w} gives correlated codes good for learning, and because binding is lossy every recovery is approximate at best. VaCoAl's \textbf{two operating modes} (Rescue vs Don't Care), realized in PyVaCoAl's DRAM-CAM, expose a different control on an \emph{exactly-invertible} substrate: in Rescue mode, collection recovery is exact (\texttt{CR1 = 1.0}, Exact Match relative to the Frontier Size); in Don't Care mode, a controlled residual irreversibility injects a small variance into \texttt{CR2} via the generational product, yielding an emergent, STDP-like pruning of deep paths --- an Occam's-razor selection with a closed-form decay, which PyVaCoAl measured at \texttt{CR2 $\approx$ 0.905} by the 56th generation.

Consider what happens at each extreme. Set the rescue circuit fully on and every block collision is repaired, so \texttt{CR1} is pinned to 1.0 at every step and \texttt{CR2 = $\Pi$\textsubscript{i} CR1(i)} is identically 1 for every candidate path. The system's retrieval is now perfect by any external standard --- and precisely because it is perfect, every reachable path is indistinguishable from every other, and the ordering among them falls back on whatever arbitrary default the implementation supplies. Set it off at sufficient depth and collisions are rare but present; \texttt{CR1} sits just below unity, varying from step to step with how many blocks were lost, and that variation, compounded down a path, is the only thing that tells one route from another.

\begin{center}\small
\begin{tabular}{@{}p{3.3cm}p{5.3cm}p{5.3cm}@{}}
\toprule
\textbf{Mode} & \textbf{Property gained} & \textbf{Property relinquished} \\
\midrule
Rescue (circuit on) & FS-relative Exact Match; zero collisions; the reflexive read of \S{}6 is exact & \texttt{CR2} discriminating power --- every path homogenized to 1.0, so no credit can be assigned \\
\addlinespace
Don't Care (circuit off) & Path-dependent credit; STDP-like pruning of detours; valuation at L3 & The zero-collision guarantee --- the reflexive read degrades gracefully along the \texttt{CR2} curve \\
\bottomrule
\end{tabular}
\end{center}

\begin{center}\small\emph{Table 3. The two modes are one circuit under one parameter. Minsky's introspective
audit (\S{}6) wants the first row; Minsky's credit assignment (\S{}7.4) wants the second.}\end{center}

\subsection{The general form}
It is easy to read Table 3 as a configuration trade-off. It is a constraint, and it is the reason this section carries a heading rather than a footnote.

The information needed to prefer one route over another is generated only as a by-product of failing along it. Credit, path quality, and any answer to \emph{which of my earlier choices deserves the reward} are all differences in how badly things went, and a system that never goes badly has no differences to report. Perfect recall and zero discrimination are not in tension here; they are the same fact stated twice. This is why the trade in Table 3 is not a defect of the present implementation to be designed away in a later version: any architecture that repairs all of its errors forfeits, by that act, the ability to rank what it retrieves --- and Minsky's engine of learning-how-to-think is exactly a ranking of one's own retrieved routes.

The consequence for the Emotion Machine reading is sharper than a design note. Minsky's Reflective layer wants an exact record of what was thought; Minsky's credit assignment, which is what makes reflection worth having, wants a record of how hard the thinking was. These are the same record read under two settings, and no setting supplies both maximally. A mind of this kind must therefore \emph{choose}, per question, whether it is auditing or valuing --- which makes the choice itself a meta-level act, and puts it exactly where \S{}10 says the missing controller belongs. We regard this as the most concrete thing this paper has to say about what a meta-control loop would be \emph{for}: not merely deciding when to introspect, but deciding which kind of memory to introspect with.

\subsection{The setting is part of the answer}
A smaller consequence follows for how results of this kind should be reported. Because the two modes yield genuinely different objects --- an exact factorization on one side, a credited path ranking on the other --- the natural output of the architecture is not one answer but a pair indexed by the setting that produced it. Reporting ``in Rescue mode the trace factorizes exactly as \texttt{(g, m, e, a)}; in Don't Care mode the direct route outranks the detour by this margin'' is strictly more informative than reporting either alone, and it is a report a system that produces one plausible continuation cannot make. The mode parameter is not noise to be tuned out before publication; it is part of what was found.

\subsection{What this is not}
Two honesty guards. First, in Tier 2, with no exact substrate to stand on, the only knob is representational geometry, forcing the learning-versus-cognition trade-off and admitting spurious factorizations; in Tier 3 the trade-off becomes a tunable choice between exact recovery and emergent selection, and when introspection is called for it can be had exactly. We do not claim to dissolve the learning-versus-cognition trade-off wholesale; we claim that on the introspection-relevant side, exactness removes the approximation and the conditionality the leading framework must live with. Second, \S{}8.2 is stated at the level of this architecture and \emph{argued}, not proved, to hold beyond it; \S{}10 gives the observation that would retire it.

\section{Two Independent Corroborations}
Two recent works reach the substrate from opposite directions. \textbf{Sutor et al. [2025]} independently arrive at the same algebra --- binding as an XOR involution, bundling as a majority vote, sequence binding as shiftable permutation --- and argue that because an AI can be observed and modeled through such algebraic statements, hyperdimensional computing is a natural substrate for metacognition: convergence on the \emph{operations}. \textbf{Poduval et al. [2026]} establish from representational geometry that cognition, decoding, and factorization demand near-orthogonal, separable codes --- the property that makes a trace recoverable --- and, in their spurious-convergence result, supply the strongest available evidence that a \emph{lossy} substrate yields only conditionally faithful recovery: convergence on the \emph{requirement}.

The pairing is worth naming for what it is. Two groups working from premises we do not share --- one from metacognitive modeling, one from representational geometry --- arrive at the operations we use and at the requirement we claim is binding. That is the signature of a constraint rather than of a taste, and it is the same form of argument the sibling papers make from biology: different starting points, same function. We borrowed the problem from Minsky; the algebra came from elsewhere, and the fact that others reached it independently is evidence that the problem admits few solutions.

Two boundaries keep the positioning honest. Sutor et al.'s metacognition is largely \emph{third-person} meta-modeling --- a hypervector model of \emph{another} system --- whereas Minsky's Reflection is \emph{first-person} introspection over one's own trace; bridging them is the meta-control problem of \S{}10. And what PyVaCoAl has \emph{demonstrated} is self-contained multi-hop deliberation with faithful traces; ingesting an external model and meta-modeling it with exact recovery is application-level future work, not a present claim.

\section{Reservations, Open Problems, Testable Predictions}
\textbf{Sufficient conditions for Reflection.} A faithful, factorizable trace is supplied; two pieces are not. First, a \textbf{meta-control loop} that decides \emph{when} to unbind and inspect the trace, and reasons about the \emph{choice of resources} rather than the object-level path --- naturally designed, we suggest, as a controller that dynamically sets the rescue mode, making introspection a layer that modulates the reversibility of the deliberation beneath it. Section 8.2 gives that suggestion its content: because audit and credit want opposite settings, the controller is not merely a scheduler but the thing that chooses which of the two the moment calls for. Second, \textbf{temporal context-tagging} that separates ``what I was thinking then'' from ``what I am thinking now.'' The architecture makes this concrete rather than mysterious: a context is just another role, so a time- or context-tagged trace is \texttt{D{\char95}t = D $\oplus$ Bind(Context, t)}, and introspection over the right window is an unbind against the matching context key. Turning that observation into a working scheme --- capacity, interference between windows, and the control that selects a window --- is the chief open problem.

\textbf{The layers above.} Self-Reflective and Self-Conscious thinking add evaluative and normative self-assessment, requiring value and affect binding. These are matters of evaluation, not introspection; the exact trace is the necessary basis on which such higher control would be built, not a substitute for it.

\textbf{Honesty guards.} The Exact Match is \emph{FS-relative} --- conditional on frontier, mode, and memory depth, not exhaustive over the combinatorial space. It is VaCoAl's architectural guarantee (collision-avoidance in Rescue mode), demonstrated at the tens-of-millions scale by PyVaCoAl's large-FS DRAM-CAM; the SRAM-CAM (CASRAM) incarnation would run the same guarantee faster and at lower power but only up to the smaller Frontier Size its capacity affords, and the full hardware measurement is future work. The architecture's exactness covers the introspection-relevant operations --- unbinding a \emph{known} role to recover its filler, and recovering collection membership --- which is exactly what introspecting one's own trace requires, since a reasoner knows the roles of its own thought; it is not a claim to solve blind multi-factor resonator factorization more accurately. Bit-exact reversibility holds in silicon; under biological noise it holds only approximately, in the statistical sense the companion papers develop. Finally, no introspection experiment is reported in this paper: the reflexive read is defined, its faithfulness is argued from the algebra, and the measurements that would confirm it are listed immediately below as predictions rather than results.

\textbf{Testable predictions.} (i) \emph{Introspective faithfulness.} A reflexive-unbind read of a deliberative bundle should recover the actually-bound fillers with an error that scales with collision saturation, not with training-data coverage --- the opposite signature from an LLM's self-report, whose faithfulness does not improve with capacity in this way. (ii) \emph{Calibrated introspective confidence.} \texttt{CR1} on the reflexive read should track the true probability of correct constituent recovery, giving a self-report whose confidence is calibrated rather than systematically overconfident. (iii) \emph{Reflection--deliberation dissociation.} Because valuation lives in Don't Care mode and audit in Rescue mode, PyVaCoAl run in Rescue mode should introspect its trace exactly while losing the emergent path-selection, and in Don't Care mode should exhibit selection while its reflexive read degrades gracefully with depth along the \texttt{CR2} curve --- a dissociation with a predicted quantitative form. (iv) \emph{A refutation we would accept.} If a retrieval system repairs every failure and nevertheless ranks its own reachable routes on a semantically contentful criterion generated internally, without an external scoring module, then \S{}8.2 is wrong and the audit/credit tension it reports is an artifact of our design. We do not think such a system exists, but the burden is ours and this is where it should be discharged.

\section{Conclusion}
Marcus told us \emph{what} an algebraic mind must do and left the substrate a conjecture; a companion paper answered the conjecture with the VaCoAl architecture --- an exact algebra over Galois fields --- and climbed Pearl's causal ladder off its reversibility. Minsky tells us \emph{how} a mind is stacked --- deliberation beneath reflection beneath self-reflection --- and supplies the second vertical axis the same reversibility unlocks. The two frameworks meet at one hinge: a representation whose structure is recoverable is, when the structure is one's own deliberative trace, an act of introspection. Reflexive unbinding of a role--filler bundle, cleaned by VaCoAl's confidence measure, is the operation; only exact reversibility makes it faithful rather than spurious. The algebraic mind Marcus described can now be built; from Minsky's viewpoint, building it is also the first rung of the climb toward a mind that can watch itself think. Deliberation is \emph{demonstrated} by PyVaCoAl; Reflection's reachability is \emph{argued}; its sufficient conditions --- meta-control and temporal context-tagging --- are named as future work; and evaluative self-assessment is left, deliberately, to the layers above.

Two consequences deserve to be carried out of this paper separately from the architecture that produced them.

The first is that verticality is not an addition. Reflection, read as introspection, required no operation the horizontal pillars had not already supplied --- only a changed argument. The corresponding negative is the one that bites: an architecture whose composition is lossy cannot acquire reflection by gaining a monitor, because exactness of inverse is a property of the elementary operation and not of what is stacked above it. A representation that cannot be exactly taken apart was never composed; it was mixed, and there is nothing in it for a monitor to read. Whether a mind can watch itself think is therefore settled at the level of how it binds, long before anything that looks like watching is built.

The second is that the ability to credit one's own routes is purchased with the ability to fail along them. Rescue mode and Don't Care mode are the same circuit under one parameter, and moving from the second to the first buys an exact audit at the cost of every route becoming indistinguishable from every other. Credit is a difference in how badly things went; a system that never goes badly has none to assign. Minsky's two demands --- see your own reasoning, and learn which of your choices deserved the reward --- therefore want the memory set two different ways, and the choice between them is itself a meta-level act. That is not a defect to be engineered away in a later version. It is, we suspect, what having a meta-level is for.

Both properties are, in the end, the same property seen from two sides: a memory that can be taken apart can tell you what it did, and a memory that keeps a record of its own difficulties can tell you which of the roads it took was worth taking.

\section*{Appendix A. A ledger of claims}
Because this paper mixes an architectural argument, a software demonstration, an unbenchmarked hardware incarnation, and a piece of philosophy of mind, we set out in one place which category carries which claim. The categories are: \textbf{demonstrated} (measured in PyVaCoAl, the large-FS DRAM-CAM incarnation); \textbf{architectural} (a property of VaCoAl's algebra or collision-avoidance, true by construction but, where it concerns speed or power, unbenchmarked here); \textbf{argued} (follows from the architecture by an argument we give, but not measured); and \textbf{deferred} (named as future work and not claimed at all).

\begin{center}\small
\begin{tabular}{@{}p{7.6cm}p{3.0cm}p{3.4cm}@{}}
\toprule
\textbf{Claim} & \textbf{Category} & \textbf{Where} \\
\midrule
\texttt{Unbind(Bind(R,F),R) = F} in the noiseless case & Architectural & \S{}2.1 \\
Recovery collision-free and exact within a Frontier Size & Architectural & \S{}2.1 \\
Multi-hop deliberation at 57 generations, 25.5M paths & Demonstrated & \S{}2.1, \S{}5 \\
\texttt{CR2 $\approx$ 0.905} at generation 56 in Don't Care mode & Demonstrated & \S{}8.1 \\
Three-tier ordering by trace faithfulness & Argued & \S{}4 \\
Reflection = reflexive unbinding of one's own trace & Argued & \S{}6 \\
The reflexive read is faithful within the Frontier Size & Argued (from architecture) & \S{}6 \\
Panalogy = content-addressable best-match retrieval & Argued & \S{}7.1 \\
One-shot, real-time, continual accumulation & Architectural & \S{}7.3 \\
Credit assignment = reflexive counterfactual simulation & Argued & \S{}7.4 \\
Audit and credit want opposite settings of one knob & Argued (from \S{}8.1) & \S{}8.2 \\
Nanosecond in-memory matching; 1/N low-power rule & Architectural, unbenchmarked & \S{}2.1, \S{}10 \\
Meta-control loop; temporal context-tagging & Deferred & \S{}10 \\
Evaluative self-assessment (L5--L6) & Deferred & \S{}5, \S{}10 \\
Introspection experiments (predictions i--iii) & Deferred & \S{}10 \\
\bottomrule
\end{tabular}
\end{center}

\begin{center}\small\emph{Table 4. Claim ledger. A reader who disagrees with a category is disagreeing with
something checkable, which is the point of printing it.}\end{center}

Two notes on the ledger. It contains no row for a biological claim, because this paper makes none; the DG--CA3 reading belongs to the sibling Perspective and is cited here only as context. And the largest single category is \emph{argued} --- which is the honest shape of a paper whose central contribution is that an existing operation, pointed in a new direction, is an act of a kind nobody had named it as.

\section*{References}
\begingroup\small\setlength{\parindent}{0pt}
[Minsky, 2006] M. Minsky. \emph{The Emotion Machine: Commonsense Thinking, Artificial Intelligence, and the Future of the Human Mind.} Simon \& Schuster.\par\smallskip
[Marcus, 2001] G. F. Marcus. \emph{The Algebraic Mind: Integrating Connectionism and Cognitive Science.} MIT Press.\par\smallskip
[Chuma et al., 2026a] H. Chuma, K. Otsuka, Y. Sato. \emph{Beyond LLMs, Sparse Distributed Memory, and Neuromorphics: A Hyper-Dimensional SRAM-CAM ``VaCoAl'' for Ultra-High Speed, Ultra-Low Power, and Low Cost.} arXiv:2604.11665.\par\smallskip
[Chuma et al., 2026b] H. Chuma, K. Otsuka, Y. Sato. \emph{How to Build Marcus's Algebraic Mind: Algebro-Deterministic Substrate over Galois Fields.} arXiv:2605.21379.\par\smallskip
[Chuma et al., 2026c] H. Chuma, K. Otsuka, Y. Sato. \emph{Bridging Silicon and the Hippocampus: Algebro-Deterministic Memory ``VaCoAl'' as a Substrate for Vector-HaSH and TEM.} arXiv:2605.15652.\par\smallskip
[Sutor et al., 2025] P. Sutor, C. Ferm\"uller, Y. Aloimonos. \emph{Hyperdimensional Computing for Metacognition.} 2nd Workshop on Metacognitive Prediction of AI Behavior (METACOG-25). \url{https://neurosymbolic.asu.edu/wp-content/uploads/sites/28/2025/04/METACOG25_HyperdimensionalComputingForMetacognition_Sutor_Fermuller_Aloimonos.pdf}\par\smallskip
[Poduval et al., 2026] P. P. Poduval, H. Errahmouni Barkam, X. Liu, S. Yun, Y. Ni, Z. Zou, N. D. Bastian, M. Imani. \emph{Optimal Hyperdimensional Representation for Learning and Cognitive Computation.} Frontiers in Artificial Intelligence 9:1690492. \url{https://www.frontiersin.org/journals/artificial-intelligence/articles/10.3389/frai.2026.1690492/full}\par\smallskip
[Frady et al., 2020] E. P. Frady, S. J. Kent, B. A. Olshausen, F. T. Sommer. \emph{Resonator Networks, 1.} Neural Computation 32:2311--2331.\par\smallskip
[Smolensky, 1990] P. Smolensky. \emph{Tensor Product Variable Binding and the Representation of Symbolic Structures in Connectionist Systems.} Artificial Intelligence 46:159--216.\par\smallskip
[Plate, 1995] T. A. Plate. \emph{Holographic Reduced Representations.} IEEE Trans. Neural Networks 6(3):623--641.\par\smallskip
[Kanerva, 1988] P. Kanerva. \emph{Sparse Distributed Memory.} MIT Press.\par\smallskip
[Kanerva, 2009] P. Kanerva. \emph{Hyperdimensional Computing: An Introduction to Computing in Distributed Representation with High-Dimensional Random Vectors.} Cognitive Computation 1:139--159.\par\smallskip
[Rahimi et al., 2018] A. Rahimi, P. Kanerva, L. Benini, J. M. Rabaey. \emph{Efficient Biosignal Processing Using Hyperdimensional Computing.} Proceedings of the IEEE 107:123--143.\par\smallskip
[Mitrokhin et al., 2019] A. Mitrokhin, P. Sutor, C. Ferm\"uller, Y. Aloimonos. \emph{Learning Sensorimotor Control with Neuromorphic Sensors: Toward Hyperdimensional Active Perception.} Science Robotics 4(30).\par\smallskip
[Pearl \& Mackenzie, 2018] J. Pearl, D. Mackenzie. \emph{The Book of Why.} Basic Books.\par\smallskip
[Turpin et al., 2023] M. Turpin, J. Michael, E. Perez, S. Bowman. \emph{Language Models Don't Always Say What They Think: Unfaithful Explanations in Chain-of-Thought Prompting.} arxiv: 2305.04388.\par\smallskip
\endgroup
\end{document}